\documentclass[11pt,a4paper]{article}

\usepackage[T1]{fontenc}
\usepackage[utf8]{inputenc}
\usepackage[english]{babel}

\usepackage{amsmath,amssymb,amsfonts,amsthm}

\usepackage{graphicx}
\usepackage{booktabs}
\usepackage{multirow}
\usepackage{subcaption}
\usepackage{xcolor}

\usepackage{geometry}
\usepackage{setspace}
\usepackage{hyperref}

\title{Proof of Concept: Multi-Target Wildfire Risk Prediction and Large Language Model Synthesis}

\author{
Nicolas Caron\thanks{Université Marie et Louis Pasteur, CNRS, FEMTO-ST, Belfort, France. \href{mailto:nicolas.caron@univ-fcomte.fr}{nicolas.caron@univ-fcomte.fr}}
\and
Hassan Noura\thanks{Université Marie et Louis Pasteur, CNRS, FEMTO-ST, Belfort, France. \href{mailto:hassan.noura@univ-fcomte.fr}{hassan.noura@univ-fcomte.fr}}
\and
Christophe Guyeux\thanks{Université Marie et Louis Pasteur, CNRS, FEMTO-ST, Belfort, France. \href{mailto:christophe.guyeux@univ-fcomte.fr}{christophe.guyeux@univ-fcomte.fr}}
\and
Benjamin Aynes\thanks{SAD Marketing, Lille, France. \href{mailto:b.aynes@sad-marketing.com}{b.aynes@sad-marketing.com}}
}

\date{}

\begin{document}
\maketitle

\begin{abstract} Current state-of-the-art approaches to wildfire risk assessment often overlook operational needs, limiting their practical value for first responders and firefighting services. Effective wildfire management requires a multi-target analysis that captures the diverse aspects of wildfire risk—including meteorological danger, ignition activity, intervention complexity, and resource mobilization—rather than relying on a single predictive indicator. In this proof of concept, we suggest to development of a hybrid framework that combines predictive models for each risk dimension with large language models (LLMs) dedicated to synthesizing heterogeneous outputs into structured, actionable reports. \end{abstract}

\section{Introduction}
Predicting forest-fire risk in France has become a strategic priority for prevention, preparedness, and rapid operational response. The 2022 season was particularly severe, with about \textbf{72,000 ha} burned in France, including unprecedented events outside the traditional Mediterranean focus (\href{https://www.georisques.gouv.fr/minformer-sur-un-risque/feu-de-foret}{Géorisques}; \href{https://www.ecologie.gouv.fr/sites/default/files/documents/02.06.2023_DP_feux-HD_en_pdf_rendu_accessible.pdf}{French Ministry press dossier}). At the European scale, 2022 was the \emph{second-worst} wildfire year since EFFIS records began in 2000 (\href{https://joint-research-centre.ec.europa.eu/jrc-news-and-updates/eu-2022-wildfire-season-was-second-worst-record-2023-05-02_en}{JRC/EFFIS}). Climate change is amplifying these hazards by lengthening the fire season, drying fuels, and extending exposure beyond historically affected regions (\href{https://meteofrance.com/le-changement-climatique/quel-climat-futur/changement-climatique-quel-impact-sur-les-feux-de-foret}{Météo-France on climate impacts}; \href{https://www.inrae.fr/actualites/dereglement-climatique-attise-risques-feux-forets}{INRAE}).

Operational products already support authorities and the public—for example, Météo-France’s \emph{Météo des Forêts}, which provides daily danger maps at the department level for the next two days with a four-tier scale (\href{https://meteofrance.com/meteo-des-forets}{Météo des Forêts}; \href{https://meteofrance.com/actualites-et-dossiers/actualites/meteo-des-forets-informer-sensibiliser-le-public-au-danger-incendie}{How it works}) at the department level. Yet to genuinely assist first responders, new statistical or AI-driven predictors must remain \emph{practice-oriented} and aligned with real operational needs. In particular, solutions should:
\begin{itemize}
  \item deliver appropriate \textbf{temporal resolution} (at least daily, ideally intra-day) to support early attack and resource pre-positioning;
  \item provide \textbf{spatial resolution} fine enough (e.g., kilometer-scale grids) to capture WUI interfaces and local wind corridors relevant to tactical decision-making;
  \item meet strict \textbf{cost and latency} constraints so inference runs in seconds to minutes on modest hardware (including near the field or offline);
  \item integrate smoothly with existing workflows and products (\textbf{interoperability}), offer basic \textbf{explainability}, and communicate \textbf{uncertainty} clearly to guide risk-informed actions.
\end{itemize}

In short, predictive value materializes only when models are \emph{operationalizable}: at the right scale, at the right time, at the right cost—and consistent with the doctrine of anticipation and rapid initial attack adopted by French fire and rescue services (\href{https://observatoire.foret.gouv.fr/themes/la-strategie-francaise-de-lutte-œcontre-les-feux-de-foret}{French Forests Observatory}).

\subsection{State-of-the-art}

From an AI research perspective, wildfire risk has inspired a wide range of applications—spanning ignition and spread prediction, dynamic danger mapping, fuel-moisture and flammability estimation, early detection from remote sensing and camera feeds, smoke and air-quality forecasting, post-fire damage assessment, and even resource allocation and response optimization—using machine learning, deep learning, and spatio-temporal modeling.

Kondylatos et al.\ \cite{WildfireDangerPrediction} release a daily wildfire-hazard dataset for Greece (2009--2021) at $\sim\!1\,\mathrm{km}^2$, integrating meteorology, land cover, and population. However, per-file sizes ($\sim$23\, GB) hinder memory and accessibility. SeaFire Cube \cite{karasante2023seasfiremultivariateearthdatacube} provides 21 years of global data (2001--2021) at 8-day, $0.25^\circ$ resolution across atmospheric, climate, vegetation, socioeconomic, and fire variables, which is suitable for seasonal prediction but too coarse for fine-scale management. Using this cube with GraphCast \cite{lam2023graphcastlearningskillfulmediumrange}, Michail et al.\ report AUPRC $0.64$ globally but only $0.20$ in Europe \cite{michail2025firecastnetearthasagraphseasonalprediction}, highlighting limited regional generalization. Mesogeos \cite{kondylatos2023mesogeosmultipurposedatasetdatadriven} spans the Mediterranean (2006--2022) at daily $\sim\!1\,\mathrm{km}^2$ with 27 variables and records of ignitions and burned areas $>30$\, ha; despite extraction tools and a leaderboard, volume remains a barrier, and excluding smaller fires removes signals like detection time and response efficiency; among LSTM, GTN, and Transformer baselines, LSTM performs best.

Across these resources, wildfire risk is typically framed as binary classification to produce risk maps, but very fine grids (e.g., $1\times1$\, km) exacerbate extreme class imbalance (e.g., only 1{,}228 positive pixels in \cite{WildfireDangerPrediction}), biasing standard metrics and leaving calibration underexplored. Models often echo known high-risk zones, with accuracy degrading at fine scales \cite{vilar2010model, https://doi.org/10.1002/eap.2316}. High memory footprints further constrain practical use, and none of the datasets encode the organization of firefighting units, even though actionable risk prediction should align with this operational structure.

LLMs are increasingly explored as decision-support tools in wildfire management due to their ability to integrate heterogeneous data sources and generate human-interpretable outputs. LLMs are increasingly positioned as orchestrators of complex wildfire intelligence systems. WildfireGPT~\cite{xie2024wildfiregpttailoredlargelanguage} exemplifies this trend by integrating climate data, historical fire records, and scientific literature retrieval to deliver targeted, explainable risk analyses for practitioners. Yet, WildfireGPT focuses on knowledge retrieval and contextual explanation based on large wildfire-related data sources, whereas it is not primarily intended for real-time structured operational risk reporting. Similarly, Li et al.~\cite{li2024cllmatemultimodalllmweather} explore multimodal LLMs for large-scale natural disaster classification, highlighting their adaptability across regions and their capacity to provide explanations alongside predictions, despite current limitations in fine-grained classification accuracy. The LLM does not perform forecasting in the strict sense. Instead, it classifies weather or disaster events based on already observed meteorological conditions. The task therefore corresponds to conditional event classification rather than temporal prediction, as no explicit future state or dynamic evolution is modeled. Agent-based LLM frameworks further extend this paradigm: Dolant et al.~\cite{dolant2025agenticllmframeworkadaptive} demonstrate how LLM-driven agents can explore probabilistic scenarios and generate actionable recommendations (e.g., evacuation, communication strategies), even in the absence of standardized evaluation metrics.

\subsection{Contributions}

This proof of concept addresses fundamental limitations of current wildfire-risk modeling and decision-support systems, which predominantly rely on single hazard or occurrence-based indicators and largely ignore operational firefighting constraints. Moreover, recent LLM-based approaches focus on knowledge retrieval or classification tasks and do not integrate predictive forecasting pipelines with structured operational report generation.

The main contributions of this study are:

\begin{itemize}
    \item \textbf{Operational reframing of wildfire risk.}  
    We demonstrate that wildfire risk is intrinsically multidimensional and cannot be meaningfully represented by a single hazard or occurrence-based indicator. By jointly analyzing meteorological danger, ignition activity, intervention duration, and resource mobilization, we show that operational wildfire risk must be treated as a multi-dimensional object.
    
    \item \textbf{A multi-target predictive wildfire-risk framework.}  
    We propose a predictive architecture that simultaneously forecasts four complementary wildfire-risk dimensions—environmental hazard, ignition pressure, intervention complexity, and logistical mobilization—providing a richer and more operationally relevant representation than single-index approaches.

    \item \textbf{A hybrid predictive--generative architecture for operational reporting.}  
    We introduce the first hybrid framework combining multi-target predictive models with a generative, multi-agent synthesis layer designed for structured, contextualized, and operationally oriented wildfire-risk report generation.
\end{itemize}

\section{Region of study}

This work was carried out in collaboration with the fire and rescue service of the French department of Alpes Maritimes (06). The data were provided by the departmental service and must remain confidential.

\subsection{The Alpes Maritimes and prediction zones}

The Alpes-Maritimes, located in southeastern France along the Mediterranean coast and bordering Italy, constitute a department characterized by a unique combination of mountainous terrain, dense forested areas, and urbanized coastal zones. This geographical configuration creates a sharp gradient between highly populated coastal cities—such as Nice, Antibes, and Cannes—and sparsely inhabited inland regions dominated by steep valleys and extensive vegetation. The department is particularly vulnerable to wildfires due to its Mediterranean climate, marked by hot, dry summers and strong winds, as well as the presence of flammable shrubland and forest ecosystems typical of the region. Human activity, especially during the tourist season, further increases ignition risk. As a result, the Alpes-Maritimes represent a critical area for wildfire prevention and operational response within the French national risk management framework.

\begin{figure*}
    \centering
    \includegraphics[width=\linewidth]{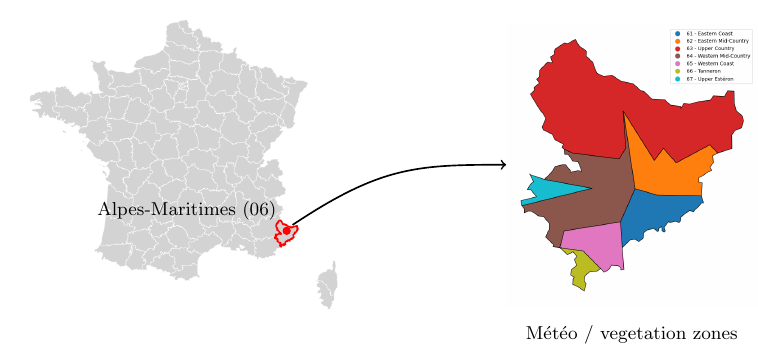}
  \caption{Position of the Alpes Maritimes in France and the segmentation in meteorological zones}
  \label{fig:region}
\end{figure*}

In this study, the department was divided into six prediction zones (see figure~\ref{fig:region}), as defined by Météo-France, each corresponding to distinct meteorological and vegetation conditions. All risk predictions and analyses were therefore conducted independently for each zone, allowing the modeling framework to account for localized environmental dynamics and spatial variability in wildfire behavior. 

The meteorological zones of the Alpes-Maritimes reflect a sharp contrast between densely populated coastal areas and the heavily forested or mountainous inland regions. Zones 61 (Eastern Coast), 65 (Western Coast), and 66 (Tanneron) correspond to highly urbanized sectors of the department, including Nice, Antibes, Cannes, and the coastal tourist corridor. These areas contain relatively little forest cover, with only limited stands of Aleppo pine and sparse Mediterranean vegetation. Their wildfire risk is therefore strongly shaped by human factors—tourism, dense infrastructure, and high seasonal population flux—rather than by fuel continuity. Despite their reduced vegetation load, these zones experience frequent ignitions linked to human activity, though fire propagation potential remains generally lower than in inland forested regions.\\

In contrast, Zones 62 (Eastern Mid-Country), 64 (Western Mid-Country), and 67 (Upper Estéron) form the forested heart of the department. These zones are characterized by extensive woodland cover and very limited urbanization. Zone 67 in particular displays a high prevalence of Scots pine, a species known for its flammability under summer drought. Zones 62 and 64 are dominated by broadleaf species, creating dense mixed forests that accumulate substantial fuel loads across the slopes and valleys of the pre-Alps. These regions present elevated fire propagation potential due to fuel continuity and topography, while their low population density reduces ignition frequency but also complicates detection and response times.\\

Finally, Zone 63 (Upper Country) corresponds to the mountainous alpine areas of the department. Comprised of high-elevation valleys and steep terrain extending toward the Mercantour massif, this zone features a large proportion of larch and coniferous forests. Its altitude, orientation, and rugged geomorphology produce distinct microclimates and strong orographic winds. Although ignition frequency is generally low due to minimal human presence, fire behavior can be severe when conditions align, with rapid spread facilitated by slope dynamics and highly flammable conifer stands. This zone also presents significant operational challenges related to access, terrain complexity, and long intervention times.

\subsection{The databases}

This study is built around two datasets obtained through collaboration with the fire and rescue services. The first dataset, covering summer period (from june to september) from 2015 to 2023, consists of a series of meteorological indices originally provided by Météo-France. This dataset enables the computation of the Danger Final Expertis\'e (DFE) risk level for each zone and each day, as defined in Section~\ref{sec:target}.\\

The second dataset contains records of wildfire interventions—including the timestamp of the first emergency call, the end-of-intervention time, and the number of engines used for natural-area fires that occurred within each zone between 2017 and 2023.

\section{A multidimensional view of wildfire risk}

In this section, we present the 4 predictive targets and demonstrate the utility of considering multiple risk values for operational needs.

\subsubsection{Targets} \label{sec:target}
The \textbf{DFE} is an operational wildfire danger index used to quantify the daily level of fire risk for each geographical zone. It is defined on an ordinal scale from 0 (low) to 4 (extreme), capturing progressively increasing levels of danger. The DFE is computed from meteorological and environmental variables provided each morning by M\'et\'eo-France, such as temperature, humidity, wind direction and speed, soil moisture indicators, and a set of indices: ICL, IH, IS, and IPP. they constitute the French adaptations of the Canadian Fire Weather Index (FWI) system. While the adaptation process is not shared by the institute, these variables jointly describe atmospheric and ecological conditions that directly influence wildfire likelihood.\\

In operational practice, the DFE serves as a central decision-support indicator for firefighting services and institutional stakeholders. Produced daily for each predefined zone, it enables localized, timely assessment of fire danger, thus informing prevention strategies, resource allocation, and situational awareness. Its ordinal structure also aligns with predictive modelling requirements, as misclassifications between adjacent categories are inherently less critical than distant ones, making it suitable for machine learning approaches to wildfire-risk forecasting.\\

A key limitation of the DFE lies in the restricted set of variables it considers. Since it is primarily driven by meteorological and vegetation-related indicators, the DFE does not account for crucial factors such as land-cover composition, population density, or calendar effects (e.g., weekends, holidays, seasonal human activity). Yet these dimensions play a significant role in shaping wildfire risk: land cover influences fuel availability and propagation potential; population patterns affect both ignition likelihood and detection speed; and calendar-related human behavior can substantially modify exposure and fire occurrence. By omitting these components, the DFE provides an incomplete representation of real-world fire danger, focusing on environmental hazard while failing to integrate human and structural dimensions that are essential for comprehensive risk assessment.
\\
The \textbf{intervention time} target corresponds to the difference, in minutes, between the first emergency call and the end of the firefighting operation. This variable has rarely been studied in the scientific literature due to the difficulty of obtaining reliable end-of-intervention timestamps; for example, several public databases such as \href{https://bdiff.agriculture.gouv.fr/}{BDIFF} (French National Forest Fire Database) do not provide the final time of intervention. Moreover, this target is highly stochastic, as it depends on factors that are difficult to represent explicitly, such as traffic conditions, road quality, and the precise location of the fires.
\\
The \textbf{Number of fires} corresponds to the total count of fire events occurring within a given zone on a specific date. In the literature, this target is often reduced to the occurrence of at least one fire, typically framed as a binary classification problem. However, such a formulation limits the expressiveness of the risk representation and can become trivial in certain contexts—for instance, during peak season or in the highest-risk areas. Similar to intervention time, this target exhibits a high degree of randomness, as fire occurrence is influenced by numerous stochastic and hard-to-model factors.\\
The \textbf{Resources} target corresponds to the total number of firefighting units (engines, vehicles, or specialized equipment) deployed across all interventions within a given zone on a given day. This variable is particularly challenging to model, as it depends not only on the characteristics of individual fires but also on operational constraints such as station-level availability, staffing, and logistical readiness. Furthermore, resource allocation practices differ across stations and may follow local doctrines or mandatory dispatch rules (e.g., minimum number of engines per call), which are difficult to encode explicitly in a predictive framework. As a result, the Resources target exhibits substantial variability that reflects organizational decisions as much as fire behavior, making it a complex and inherently noisy variable for data-driven modeling.
\\
\textbf{Time intervention}, \textbf{Number of Fires}, and \textbf{Resources} have been transformed in order to get an output similar to \textbf{DFE}. Each risk forecasting is treated as an ordinal multi-class task. A five-level occurrence label is created: days with no fires are placed in class 0, while all positive instances are grouped by K-means (which shows a good clustering performance in~\cite{caron2026extremevalueforestpredictionstudy} using BDIFF) into four ordered categories—Normal, Medium, High, and Extreme. This scheme emphasizes typical fire-activity patterns instead of relying on absolute counts.

The datasets are split into training (2015/2017-2021), validation (2022), and testing (2023) subsets. All processing was performed on the training set and then generalized to all data. Each row is composed of the aggregated features and the target risk for a specific zone on a specific day. All prepossessing calculations have been done on the training set and applied to others.

\subsection{Analysis of target distributions across meteorological zones}

\begin{figure*}[h!]
    \centering

    \begin{subfigure}{0.95\textwidth}
        \centering
        \includegraphics[width=\linewidth]{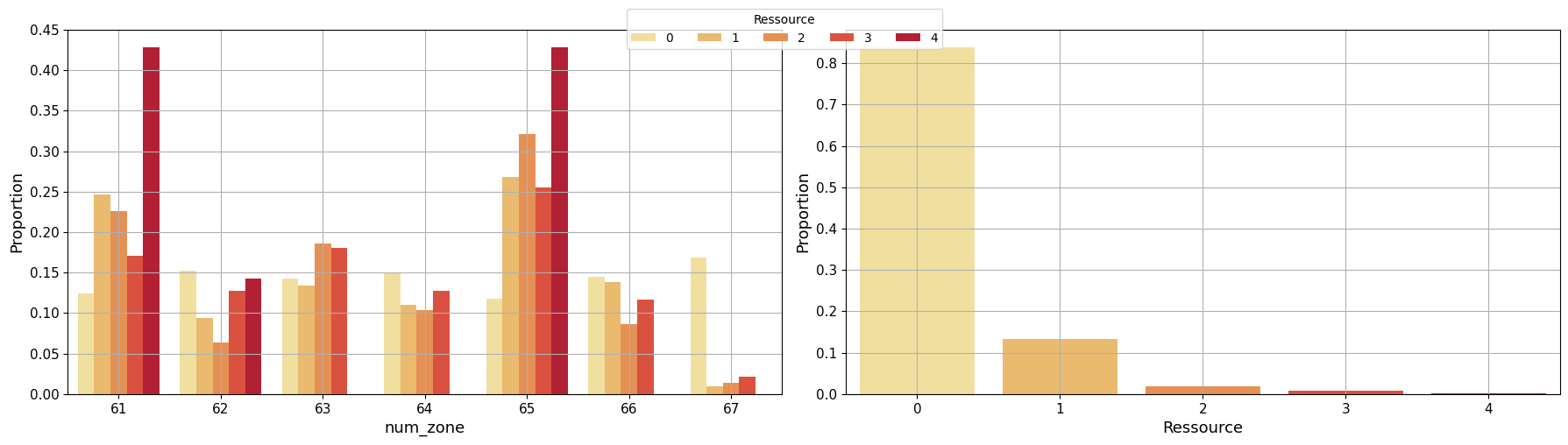}
        \caption{Normalized distribution of Resources per zone and global distribution.}
    \end{subfigure}

    \begin{subfigure}{0.95\textwidth}
        \centering
        \includegraphics[width=\linewidth]{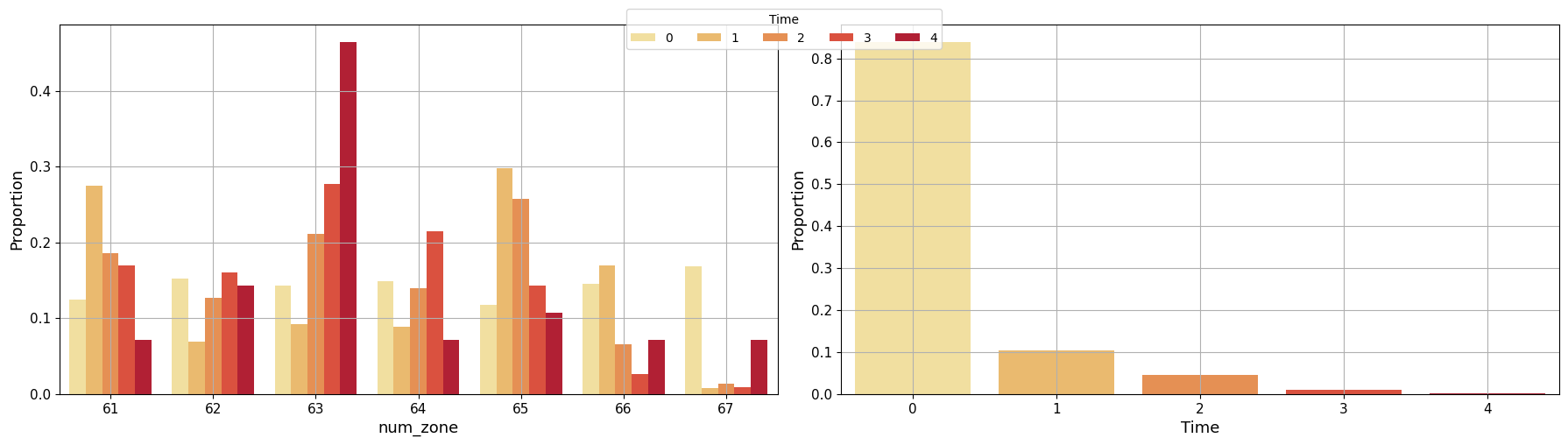}
        \caption{Normalized distribution of Intervention Time per zone and global distribution.}
    \end{subfigure}

    \begin{subfigure}{0.95\textwidth}
        \centering
        \includegraphics[width=\linewidth]{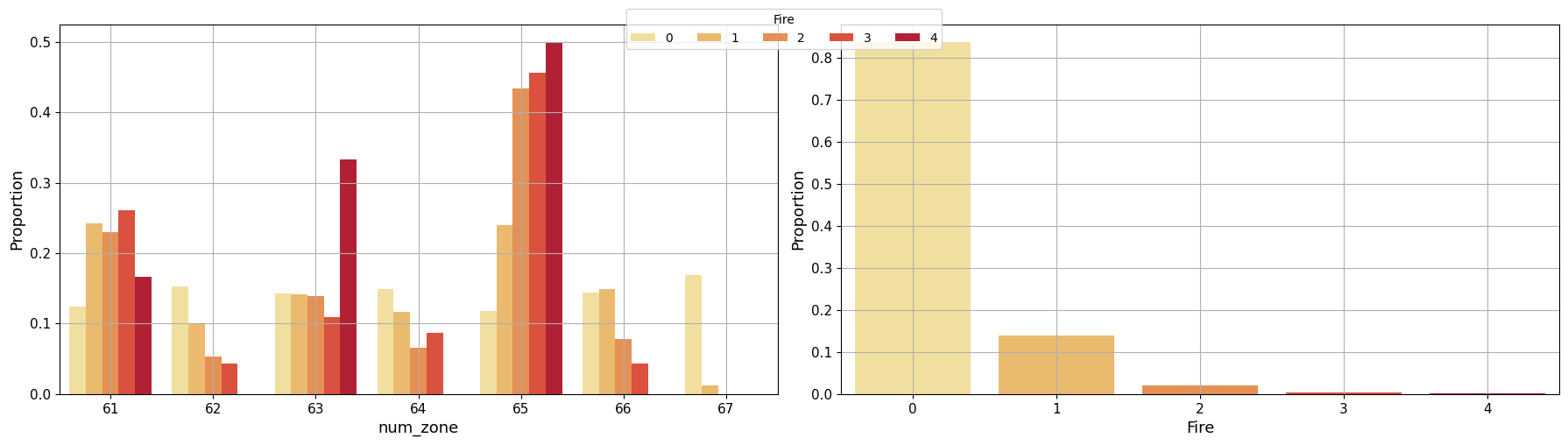}
        \caption{Normalized distribution of Number of Fires per zone and global distribution.}
    \end{subfigure}

    \begin{subfigure}{0.95\textwidth}
        \centering
        \includegraphics[width=\linewidth]{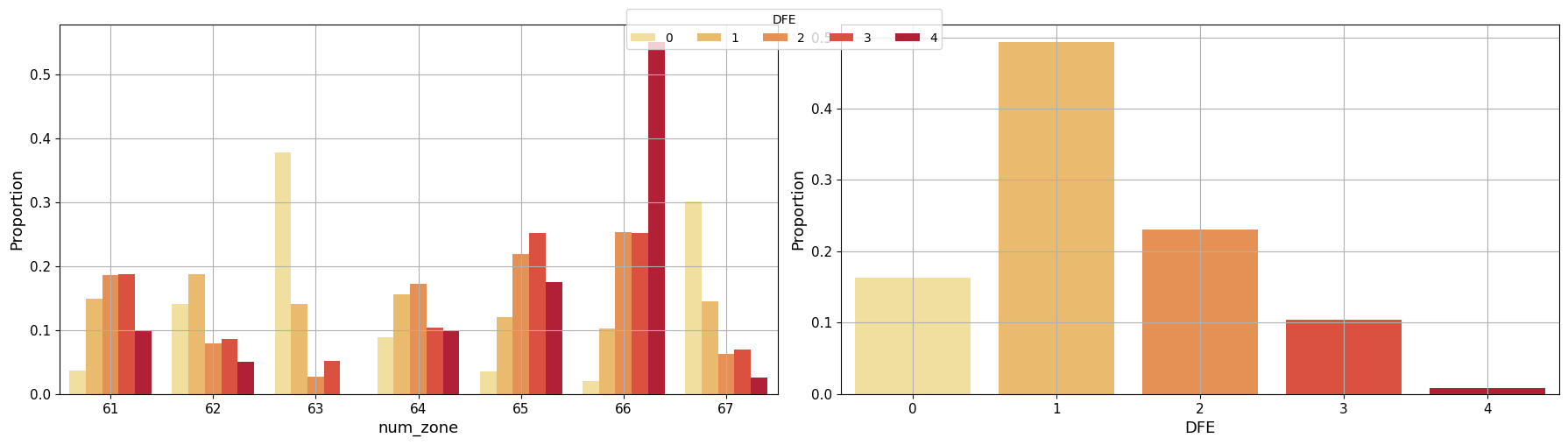}
        \caption{Normalized distribution of DFE per zone and global distribution.}
    \end{subfigure}

    \caption{Normalized distributions of the four predictive targets across the seven meteorological zones.}
    \label{fig:target-histograms}
\end{figure*}

To better understand the relationships between the different predictive targets, we examine their normalized distributions across the seven meteorological zones (\texttt{61--67}) defined by Météo-France. Figures~\ref{fig:target-histograms} present, for each target, two complementary histograms: (i) the distribution of the target per zone and (ii) its global distribution. These visualizations allow us to assess both spatial heterogeneity and overall imbalance.

\subsection{Interpretation and correlation between targets}

\paragraph{Zone \texttt{61}.}  
The distributions of Resources and Number of Fires exhibit appreciable activity, driven by strong human presence and tourist pressure. However, Intervention Time remains dominated by short-duration events, indicating efficient access and rapid response. The DFE distribution, although moderate compared with forested zones, still displays a non-negligible proportion of elevated classes during peak-season meteorological conditions.

\paragraph{Zone \texttt{62}.}  
As a more rural and forested zone \texttt{62} shows lower ignition frequencies and a distribution heavily dominated by zero-fire days. Resources and Intervention Time remain concentrated in the lowest classes, reflecting limited operational demand. Elevated DFE classes appear, but with low intensity and frequency.

\paragraph{Zone \texttt{63}.}  
This mountainous zone shows a distinct behaviour: while the resources per day remain low, Intervention Time and number of fires display heavier tails toward higher classes. This can suggest that most of the fires are low-risk (not very dangerous). The high intervention time can be explained by the difficult accessibility to the ignition point. The DFE distribution includes a higher proportion of low classes, reflecting the influence of altitude.

\paragraph{Zone \texttt{64}.}  
Predominantly rural and forested, \texttt{64} exhibits moderate ignition activity and distributions dominated by low-resource and moderate intervention time. Nevertheless, the DFE distribution shows a significant presence of intermediate and high classes (similar distribution to Zone\texttt{61}).

\paragraph{Zone \texttt{65}.}  
This western coastal zone shows patterns similar to \texttt{61}, with more frequent fires and regular operational mobilization. The distributions of Resources and Number of Fires feature higher activity, consistent with dense human presence. The DFE distribution tends to be relatively higher than in the other zones, with more frequent elevated classes. This tends to confirm the deep correlation between population density and fire intervention.

\paragraph{Zone \texttt{66}.}  
Despite being urban and only weakly forested, \texttt{66} presents surprisingly high proportions of elevated DFE classes. This contrast between operational activity (relatively modest in terms of fires, resources, and intervention times) and meteorological hazard (frequently assessed as high) reflects limitations inherent to the DFE index. Its construction emphasizes atmospheric and surface dryness rather than vegetation continuity or actual ignition pressure. This decoupling illustrates that DFE may overestimate local hazard in urban areas, where ignition probability and fire spread potential remain structurally lower.

\paragraph{Zone \texttt{67}.} 
This heavily forested inland zone is characterized by low ignition frequency and operational activity concentrated in the smallest classes of Resources and Intervention Time. The DFE distribution shows a consistent presence of low classes.

\begin{figure}
    \centering
    \includegraphics[width=0.7\linewidth]{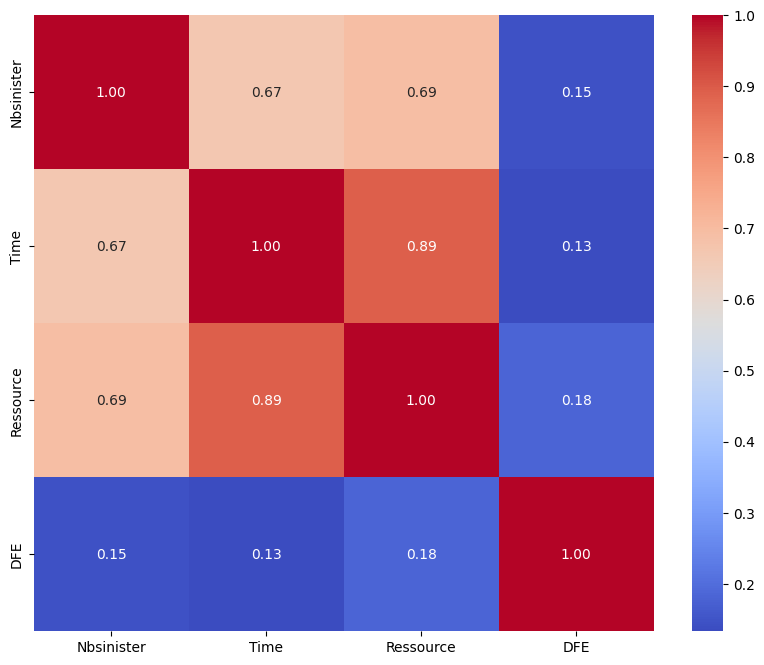}
    \caption{Correlation matrice between each targets.}
    \label{fig:correlationMatrice}
\end{figure}

Figure~\ref{fig:correlationMatrice} clearly shows that the different risk values exhibit distinct correlation patterns. The targets related to the number of fires, resources, and response time are highly correlated (particularly resources and response time), which is expected since they originate from the same intervention dataset. We observe that the correlation decreases as the number of incidents increases, indicating that the relationship between the number of fires and extreme interventions does not necessarily hold. Finally, correlations with the DFE risk are relatively weak. This can be explained by the difference in signal types (continuous for DFE and discrete for interventions). Consequently, the DFE risk alone does not provide a sufficiently informative risk value to explain the other target variables. Moreover, because intervention data are discrete (intervention or not), they do not represent a continuous risk, which may significantly limit interpretability.

\section{Training methodology}

This section presents the methodology employed in this article to predict the four presented targets.

\subsection{Features}

The features associated with the DFE risk are provided directly in the original dataset, whereas variables such as intervention time, resources, and the number of fires tend to depend on additional factors, such as population density, land cover, and calendar information (holiday, weekend, etc). Therefore, for these two targets, we used supplementary explanatory variables using a database proposed by Caron et al.~\cite{caron2025localizedforestriskprediction} to better capture their underlying drivers.

The features computed for training the models are grouped into six categories, as shown in Table~\ref{tab:variables}: Meteorological, Topographic, Socio-Economic, and Historical. Feature processing is the same method as in the original article. The integration follows the same scheme: features are transformed into a 3D raster with a resolution of 2 km, aggregated using average, maximum, and minimum. Features with low variance or high correlation (measured with Pearson, Spearman, and Kendall methods, keeping those with the highest variance) are removed.

\begin{table*}[htbp]
    \centering
    \caption{Summary of features used in this study. ``--'' indicates the same as above.}
    \label{tab:variables}
    \small
    \resizebox{\textwidth}{!}{%
    \begin{tabular}{p{4.2cm}p{2.1cm}p{3.0cm}p{4.2cm}p{2.1cm}p{3.0cm}}
        \toprule
        Variables & Frequency & Source & Variables & Frequency & Source \\
        \midrule
        \multicolumn{3}{l}{\textbf{Meteorological}} & \multicolumn{3}{l}{\textbf{Topographic}} \\
        \midrule
        Temperature                          & 12h, 16h & Meteostat                & Elevation                        & Static & IGN (CourbesDeNiveau) \\
        Dew Point                            & --       & --                       & Forest land cover                & --     & IGN (BD Forêt)         \\
        Precipitation                        & --       & --                       & Land cover                       & --     & Corine                 \\
        Wind direction                       & --       & --                       & NDVI, NDSI, NDMI, NDBI, NDWI     & 8 days & Landsat (GEE)         \\
        Wind speed                           & --       & --                       &      & Static & --                     \\
        Precipitation in last 24 hours       & --       & --                       &                                  &        &                        \\
        Snow height                          & --       & --                       &                                  &        &                        \\
        Sum of last 7 days rainfall          & --       & --                       &                                  &        &                        \\
        Days since last rain                 & 12h      & --                       &                                  &        &                        \\
        Nesterov                             & --       & Firedanger               &                                  &        &                        \\
        Munger                               & --       & --                       &                                  &        &                        \\
        KBDI                                 & --       & --                       &                                  &        &                        \\
        Angstroem                            & --       & --                       &                                  &        &                        \\
        BUI, ISI, FFMC, DMC, FWI            & --       & --                       &                                  &        &                        \\
        Daily Severity Rating                & --       & --                       &                                  &        &                        \\
        Precipitation index (3, 5, 9 days)   & --       & Calculated               &                                  &        &                        \\
        \midrule
        \multicolumn{3}{l}{\textbf{Socio-economic}} & \multicolumn{3}{l}{\textbf{Historical}} \\
        \midrule
        Highway                              & Static   & IGN (BD Route)           & Past risk                        & Daily  & Calculated             \\
        Population                           & --       & Kontur                   &          &      &                      \\
        Calendar                             & Daily    & --                       & Cluster                          & Static & --                     \\
                                             &          &                          & Department                       & Static & --                     \\
        \bottomrule
    \end{tabular}%
    }
\end{table*}

\begin{figure*}
    \centering
     \centerline{\includegraphics[width=1.2\columnwidth]{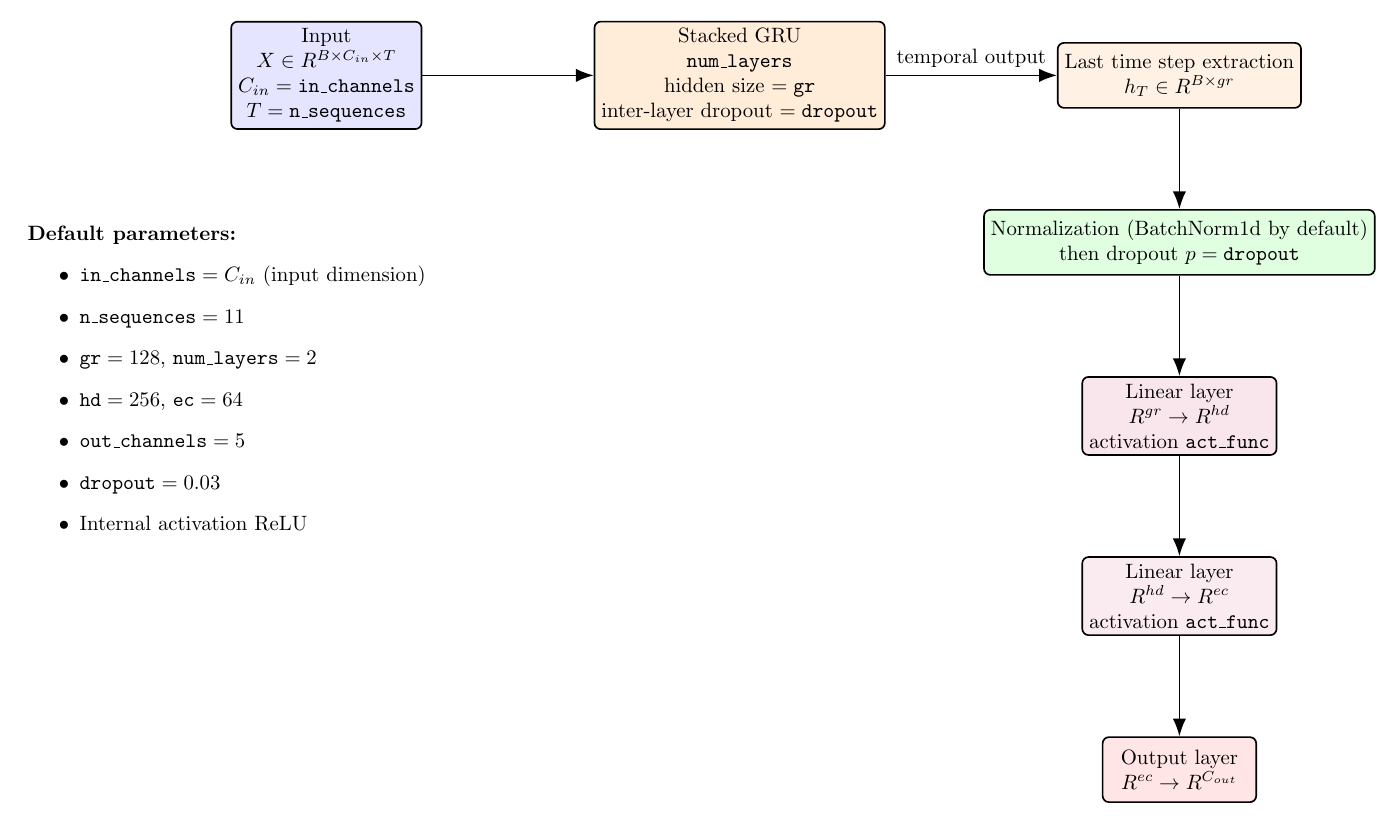}}
    \caption{GRU neural network used in this article.}
    \label{fig:gru}
\end{figure*}

We consider a stacked Gated Recurrent Unit (GRU) (figure~\ref{fig:gru}) network that processes 
an input tensor $X \in \mathbb{R}^{B \times C_{\mathrm{in}} \times T}$, 
where $B$ denotes the batch size, $C_{\mathrm{in}}$ the number of input channels, 
and $T$ the number of temporal sequences. 
The GRU consists of a configurable number of stacked layers 
(\texttt{num\_layers}), each with hidden size $g_r$, 
and optional inter-layer dropout with probability $p = \texttt{dropout}$. 
From the final GRU layer, only the hidden state at the last time step 
$h_T \in \mathbb{R}^{B \times g_r}$ is extracted as the temporal output. 
This output is normalized using batch normalization, followed by dropout with rate $p$. 
A fully connected layer then maps 
$\mathbb{R}^{g_r} \rightarrow \mathbb{R}^{h_d}$, 
followed by an activation function $\texttt{act\_func}$ (e.g., ReLU). Then another linear layer 
$\mathbb{R}^{h_d} \rightarrow \mathbb{R}^{e_c}$ 
with the same activation. 
Finally, an output layer maps 
$\mathbb{R}^{e_c} \rightarrow \mathbb{R}^{C_{\mathrm{out}}}$. 
Default hyper-parameters are :
$T = 11$, $g_r = 128$, $\texttt{num\_layers} = 2$, 
$h_d = 256$, $e_c = 64$, $C_{\mathrm{out}} = 5$, 
and $\texttt{dropout} = 0.03$, 
with ReLU used as the internal activation function.

Training was performed over 3000 epochs with a patience of 100, using the WKLoss loss function and a learning rate of 0.0005.

In addition to the persistence baseline---which returns, for each zone, the last observed value at time $t-1$---we introduce three complementary reference models to contextualize predictive performance: (i) an \emph{Always-WeekOfYearZone} baseline that assigns, for each zone and ISO week of the year, the most frequent historical class, capturing seasonal operational regularities; (ii) a zone-wise Poisson baseline that models event intensity as a rare-event process and derives class predictions from the resulting rate estimates; and (iii) a multiclass Logistic Regression baseline, with imbalance handling tuned on the validation set by scanning the effective proportion of the majority class and selecting the configuration that maximizes IoU. Together, these baselines provide deterministic, seasonal, generative, and discriminative reference points against which the proposed models are evaluated.

\subsection{Under sampling}
As shown above, the class distributions for the \textit{intervention time}, \textit{resources}, and \textit{number of fires} targets exhibit a strong imbalance dominated by class~0. To improve model convergence for each of these targets, we tested several under-sampling rates ranging from 0.05 to 1.0, selecting for each case the proportion that yielded the highest Intersection over Union (IoU) score on the validation set. No modification of the class distribution was applied to the \textit{DFE} target, as its ordinal structure and more balanced spread across classes already provide suitable conditions for model training. The same 0-sample proportion has been applied on each horizon.

\subsection{Performance}

Table~\ref{table:iou} compares the Intersection over Union (IoU) scores obtained by the GRU model and several baseline methods for the four wildfire-related targets considered in this study. IoU is well suited for ordinal multi-class wildfire prediction, as it preserves class ordering and accounts for uncertainty: misclassifying a nearby risk level is penalized less than confusing distant ordinal classes~\cite{ai6100253}.

The results reveal pronounced disparities in predictability across the four targets, reflecting their fundamentally different underlying dynamics. The DFE is by far the most predictable variable, with the GRU achieving an IoU of 0.91, substantially outperforming persistence and probabilistic baselines. This confirms that meteorological hazard is governed by smooth and largely deterministic atmospheric processes that can be effectively captured by data-driven models.

In contrast, the three operational targets exhibit markedly lower predictive performance across all evaluated methods. The number of fires, resource mobilization, and intervention time reach IoU values of only 0.22, 0.22, and 0.19 respectively with the GRU model, while baseline approaches remain even lower. These consistently reduced scores highlight the intrinsically stochastic, human-driven, and organizationally constrained nature of operational wildfire dynamics. Although persistence, Poisson, and Logistic Regression baselines capture coarse seasonal and historical regularities, their limited performance confirms that operational response variables cannot be reliably inferred from environmental predictors alone.

The strong disparity between the DFE and intervention-related targets demonstrates that wildfire risk cannot be characterized by a single homogeneous variable, thereby supporting the need for a multidimensional risk representation that jointly accounts for environmental hazard and operational response indicators.

\begin{table}[h!]
\centering
\caption{Comparison between scores IoU of GRU model and Persistence model for each target. LR stands for Logistic Regression}
\begin{tabular}{l c c c c c}
\hline
\textbf{Target} & \textbf{GRU} & \textbf{Persistence} & \textbf{ALWAYS-MAX} & \textbf{Poisson} & \textbf{LR} \\
\hline
DFE & 0.91 & 0.80 & 0.46 & 0.77 & 0.89\\
Number of fires & 0.22 & 0.13 & 0.10 & 0.17 & 0.21\\
Ressource & 0.22 & 0.14 & 0.13 & 0.18 & 0.20 \\
Time intervention & 0.19 & 0.13 & 0.06 & 0.10 & 0.18\\
\hline
\end{tabular}
\label{table:iou}
\end{table}

\begin{figure*}
    \centering
     \centerline{\includegraphics[width=1.2\columnwidth]{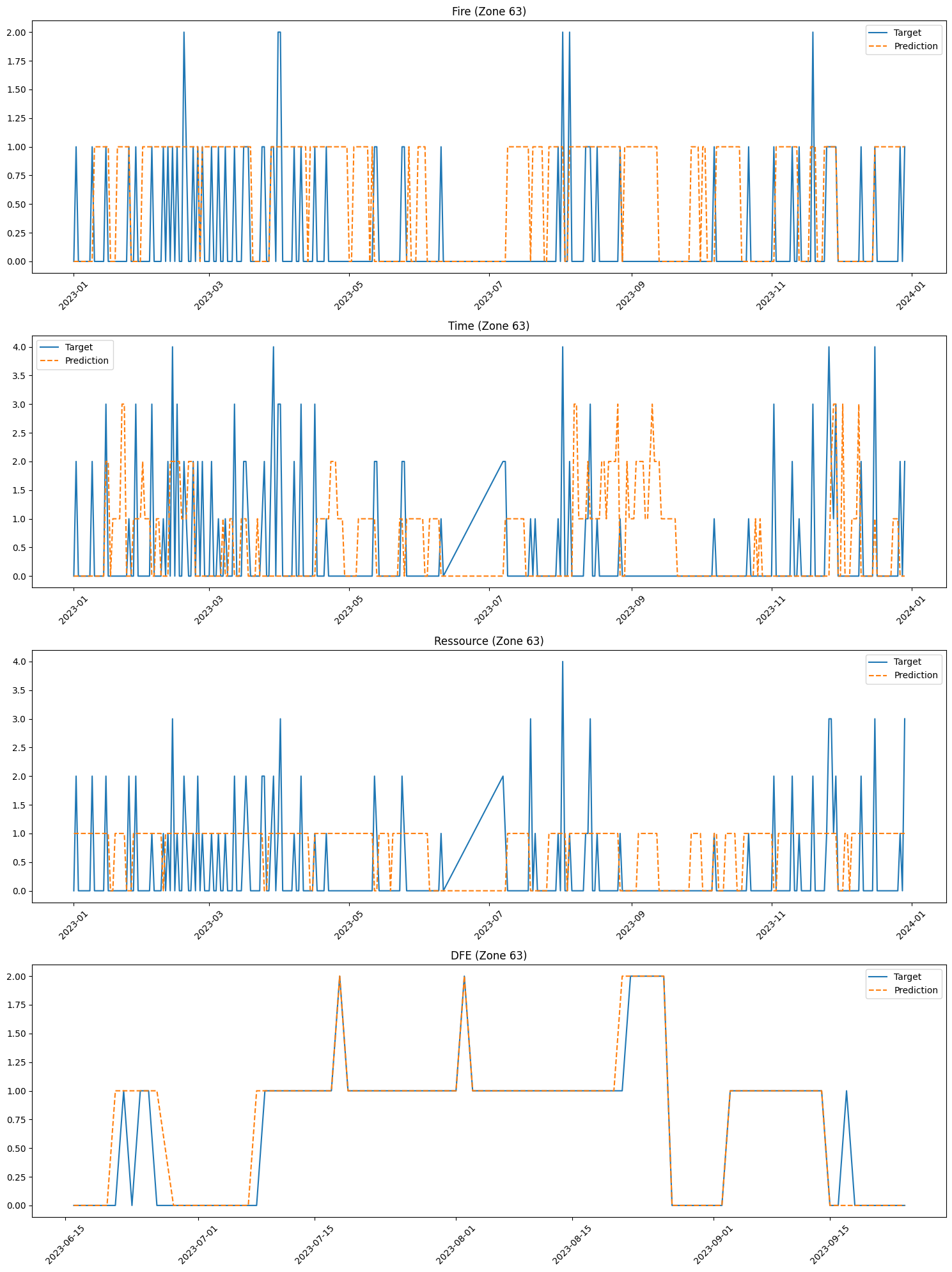}}
    \caption{Comparison of predicted and real signal in the zone 63 obtained for each target (GRU model).}
    \label{fig:res}
\end{figure*}

Figure~\ref{fig:res} shows an example of the predictions obtained for Zone 63 on the test set.

\subsection{Justification of performance differences.}
These strong discrepancies arise naturally from the fundamentally different statistical and
operational nature of the targets.

\subparagraph{1. Deterministic and feature-aligned structure of the DFE target.}
The DFE is an operational wildfire-danger index built directly from meteorological and
vegetation indicators such as temperature, humidity, wind, and FWI-derived sub-components.
These variables exhibit strong temporal autocorrelation and evolve smoothly from day to day.
Moreover, they correspond precisely to the predictors used by the model, ensuring a close
alignment between input features and the underlying causal drivers of the target. As a result,
the model can anticipate the evolution of DFE several days in advance, leading to high IoU
scores and a predictable, monotonic degradation as the horizon increases.

\subparagraph{2. Stochastic, human-driven nature of Number of Fires, Intervention Time, and Resources.}
The other three targets represent operational outcomes rather than physical or meteorological states. As described in Section~III-A of the study:
\begin{itemize}
    \item \textit{Number of Fires} depends on ignition opportunities, human activity levels, and rare-event stochasticity;
    \item \textit{Intervention Time} depends on accessibility, traffic, terrain, and precise fire locations;
    \item \textit{Resources} is driven by station availability, local dispatch rules, and organisational constraints.
\end{itemize}
These factors are not only highly variable and weakly auto-correlated, but are also only partially
captured---or not captured at all---by the available feature set. Consequently, the predictive
models lack the necessary information to reproduce or anticipate the behaviour of these
variables, yielding low and horizon-invariant IoU curves. Building a comprehensive database capable of reliably capturing these operational variables presents major practical and methodological challenges. Many of the underlying factors—such as traffic conditions, road accessibility, precise ignition locations, station-level availability, and local dispatch doctrines—are highly dynamic, heterogeneous, and often undocumented in centralized information systems. These elements are typically managed by independent operational units using heterogeneous reporting standards, and are rarely archived with sufficient temporal resolution to support predictive modeling. Moreover, several of these drivers depend on real-time human decision-making and organizational constraints that are difficult to formalize, standardize, and legally share across jurisdictions. As a result, constructing a consistent, real-time, and spatially localized database that accurately represents these operational determinants would require substantial institutional coordination, data governance agreements, and technical infrastructures, making such a dataset costly, complex, and only partially attainable in practice.. However the consistently low predictive performance observed is not necessary interpreted as a failure of the learning architecture, but as empirical evidence of the intrinsically stochastic, human-driven, and weakly observable nature of operational wildfire dynamics.

\paragraph{Implications for a Multi-variable Risk-Management Framework.}

The correlation and modeling analyses show that the four targets describe distinct and complementary dimensions of wildfire dynamics. The DFE characterizes environmental hazard driven by climatic and vegetation conditions, the number of fires reflects ignition activity, intervention time captures accessibility and operational constraints, and the resources target represents logistical mobilization. Their weak mutual correlations indicate that none of these variables alone provides a comprehensive representation of wildfire risk, which inherently combines environmental, human, and organizational components.

In addition, the modeling results reveal a strong predictability gap between meteorological hazard and intervention-related signals. The DFE can be forecast with high accuracy due to its smooth temporal dynamics and its close alignment with meteorological predictors, whereas the number of fires, intervention time, and resource mobilization exhibit weak and horizon-invariant predictability, reflecting the intrinsically stochastic and human-driven nature of operational wildfire dynamics.

A purely meteorological indicator such as the DFE may overestimate hazard in highly urbanized zones and underestimate risk in areas where ignition pressure or operational constraints dominate. Although meteorological conditions remain fundamental drivers of potential fire spread, intervention-related variables are discrete, highly stochastic, and dominated by zero-event days. Importantly, a day without recorded interventions does not imply the absence of wildfire risk, as ignitions are influenced by rare but impactful stochastic events. The limited predictability of intervention-related variables implies that raw model outputs alone are insufficient to support operational decision-making. Post-processing and contextual correction mechanisms are required to interpret, qualify, and operationally adjust predictive signals. Such post-model analysis enables the integration of domain knowledge, contextual information, and known structural blind spots of predictive models, thereby improving the robustness and operational relevance of wildfire-risk assessments.

These findings demonstrate that wildfire risk cannot be meaningfully reduced to a single scalar indicator. While the DFE provides a robust estimation of environmental hazard, it does not capture the human and organizational dimensions that shape real fire occurrence and response capacity. Conversely, operational targets—despite their low individual predictability—encode essential information related to ignition pressure, logistical demand, and intervention complexity that is absent from meteorological indices. Capturing all the determinants required for accurate prediction of intervention times and resource mobilization is inherently difficult, as several key drivers depend on real-time human decision-making and organizational constraints that are weakly observable, difficult to formalize, and only partially accessible in practice.

This complementarity motivates a multi-variable risk-management paradigm in which the joint interpretation of heterogeneous targets yields a more operationally relevant risk signal. For instance, high meteorological hazard combined with low expected resource demand may indicate environmental vulnerability without immediate operational pressure, whereas moderate hazard associated with elevated ignition or intervention indicators may reveal human-driven risk scenarios not detectable through meteorological indices alone. Jointly analyzing these dimensions enables a more comprehensive and adaptive representation of wildfire risk to support both environmental monitoring and resource planning.

Because the four targets represent fundamentally different physical, stochastic, and organizational processes, no mathematically well-posed scalar aggregation function can preserve their operational semantics simultaneously. Any scalar fusion necessarily induces an irreversible loss of meaning.

\section{Generative AI}

An in-depth analysis of correlations during training revealed that an approach based solely on a meteorological index, regardless of its robustness, is not sufficient to adequately meet the operational needs of emergency services. Indeed, intervention-related signals—such as the number of fires, crew engagement time, and resource mobilization—are particularly difficult to predict reliably due to the high level of stochasticity inherent in wildfire events.

Given these limitations, fusing multiple dimensions of risk becomes essential. Combining meteorological, operational, and contextual indicators provides a more comprehensive view, including expected fire occurrence, intervention duration, resource availability, and drought conditions.

Beyond raw prediction, the interpretability of model outputs is critical—especially to justify very high risk levels to decision-makers and to strengthen trust in AI-based systems. Fine-grained and contextualized explainability increases the credibility of alerts and facilitates their adoption in operational settings.

\begin{figure*}[ht]
    \centering
    \includegraphics[width=\linewidth]{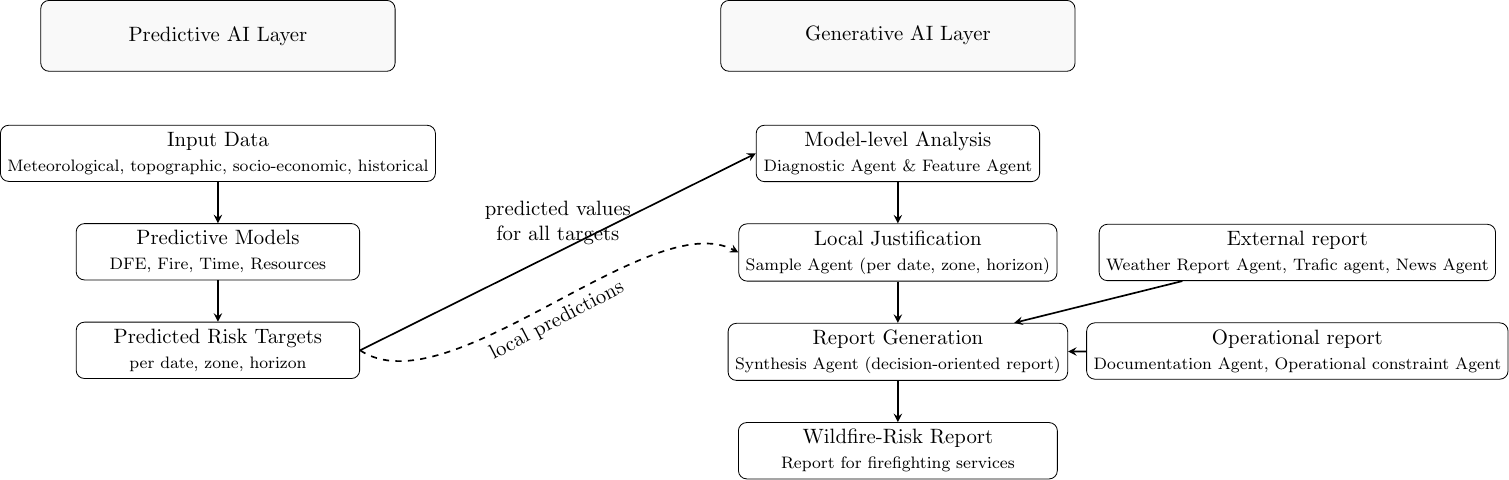}
    \caption{High-level architecture of the multi-agent framework. The Predictive AI layer forecasts multiple wildfire risk targets, while the Generative AI layer analyses, explains, and synthesizes these predictions into an operational report.}
    \label{fig:global}
\end{figure*}

We propose here a first prototype of a hybrid predictive–generative system aimed at producing a robustness-oriented and interpretability-focused wildfire risk reports that are better aligned with operational constraints than existing solutions. The system is based on a multi-layer architecture (see Figure~\ref{fig:global}): each risk dimension (hazard score, deviation from observations, importance of explanatory variables) is independently analyzed by specialized agents. Their outputs are then fused by a generative layer that synthesizes all information into a single operational report. This final report qualifies the risk level, justifies the decision, and formulates tailored recommendations, thereby providing reliable and transparent decision support for wildfire response services.

The generative layer does not statistically correct predictive models in a probabilistic sense. Instead, it performs a contextual post-model synthesis aimed at mitigating known structural blind spots of individual predictors. These blind spots originate from the intrinsically stochastic, human-driven, and organizational determinants of wildfire dynamics that are weakly observable or absent from predictive datasets. The generative layer therefore acts as an interpretability- and robustness-oriented synthesis mechanism rather than a formal error-correction module.

This multi-agent architecture could serve several key purposes:
\begin{itemize}
    \item \textbf{Synthesis:} The LLM-based framework receives predictions from all target-specific models and combines them into a single, unified wildfire risk level, tailored for operational use.
    \item \textbf{Interpretability:} Specialized agents generate human-readable justifications, reliability assessments, and operational recommendations, transforming raw model outputs into transparent, actionable reports.
    \item \textbf{Contextualization:} The system integrates additional information (such as weather reports, feature importances, and historical diagnostics), enabling nuanced risk qualification and uncertainty quantification.
    \item \textbf{Blind-spot mitigation}: Structural blind spots of individual predictive models are mitigated through contextual synthesis and domain knowledge integration.
    \item \textbf{Operational Constraint Integration:} The framework enables the incorporation of decentralized and non-centralized operational knowledge—such as local organizational structures, dispatch doctrines, and mandatory intervention rules (e.g., minimum number of units or personnel per call)—which are difficult to formalize and integrate into predictive models, but critically shape real-world firefighting constraints and response behavior.
    \item \textbf{External Data Enrichment:} The system supports the integration of external and real-time information sources—such as traffic conditions, detailed weather reports, and major ongoing events—which are not natively included in predictive datasets but significantly influence detection times, accessibility, and operational pressure during wildfire interventions.
\end{itemize}

\section{Conclusion}
Operational needs specific to firefighting services remain largely unaddressed in the current literature on wildfire risk assessment. Existing approaches tend to overlook the practical realities faced by first responders, notably the necessity for actionable, context-aware reports that synthesize multiple dimensions of risk. Our analysis demonstrates that adopting a multi-faceted view—integrating indicators such as meteorological danger, expected ignitions, intervention complexity, and resource mobilization—is essential for producing operationally relevant decision-support tools.

While LLMs have recently shown promise as flexible orchestrators of complex information, their application to the generation of field-ready wildfire risk reports remains largely unexplored. Most state-of-the-art systems either use LLMs as direct predictors or limit their scope to explanation tasks without rigorous integration with predictive models.

In this work, we propose that the combination of Predictive AI and Generative AI represents a concrete advance: predictive models quantify multi-dimensional risk, while generative agents synthesize, interpret, and communicate these findings in a form directly usable by operational teams. While this proof of concept does not claim formal reliability or certification-level guarantees but introduces a robustness-oriented architectural hypothesis for operational decision support. This hybrid architecture not only addresses critical gaps highlighted by practitioners but also paves the way for more trustworthy, explainable, and actionable wildfire risk management frameworks.

We acknowledge that publicly obtaining detailed data on firefighting resources and intervention times is challenging due to institutional, legal, and organizational constraints. However, these difficulties should not hinder research efforts in wildfire risk modeling. This work also aims to encourage closer collaboration between the academic community and firefighting services, as such partnerships are essential for developing operationally meaningful and impactful wildfire decision-support systems.

\bibliographystyle{unsrt}  
\bibliography{main}
\end{document}